\documentclass[conference]{IEEEtran}
\IEEEoverridecommandlockouts
\usepackage{cite}
\usepackage{amsmath,amssymb,amsfonts}
\usepackage{algorithmic}
\usepackage{graphicx}
\usepackage{textcomp}
\usepackage{xcolor}
\def\BibTeX{{\rm B\kern-.05em{\sc i\kern-.025em b}\kern-.08em
    T\kern-.1667em\lower.7ex\hbox{E}\kern-.125emX}}
\begin{document}

\title{Beyond Win Rates: A Clustering-Based Approach to Character Balance Analysis in Team-Based Games}
\author{Haokun Zhou}

\maketitle

\begin{abstract}
Character diversity in competitive games, while enriching gameplay, often introduces balance challenges that can negatively impact player experience and strategic depth. Traditional balance assessments rely on aggregate metrics like win rates and pick rates, which offer limited insight into the intricate dynamics of team-based games and nuanced character roles. This paper proposes a novel clustering-based methodology to analyze character balance, leveraging in-game data from Valorant to account for team composition influences and reveal latent character roles. By applying hierarchical agglomerative clustering with Jensen-Shannon Divergence to professional match data from the Valorant Champions Tour 2022, our approach identifies distinct clusters of agents exhibiting similar co-occurrence patterns within team compositions. This method not only complements existing quantitative metrics but also provides a more holistic and interpretable perspective on character synergies and potential imbalances, offering game developers a valuable tool for informed and context-aware balance adjustments.
\end{abstract}


\section{Introduction}
Competitive, team-based games with diverse character rosters, such as \textit{Overwatch}, \textit{Rainbow Six Siege}, and \textit{League of Legends}, offer rich strategic gameplay, but this very diversity presents a significant challenge: maintaining game balance. Imbalances, where certain characters are significantly stronger or weaker than others, can lead to skewed character selection rates, reduced strategic variety, and a diminished player experience. In these games, character abilities define their in-game roles and heavily influence the flow of battle. Characters with excessively unique \textit{and} overly powerful traits can become mandatory choices, limiting player agency and strategic options. Conversely, characters with irrelevant or redundant abilities can become functionally obsolete. Similarly, a high degree of overlap in character traits diminishes the strategic diversity available to players. While a universally accepted definition of game balance remains elusive, as noted by Pfau and Seif El-Nasr \cite{pfau2022game}, meaningful game diversity is considered a key indicator for understanding and quantifying game balance \cite{jaffe2013understanding}. Sirlin claims that a reasonably large number of viable options available to players, especially for high-level players, might indicate a balanced game \cite{sirlin2001balancing}. Therefore, identifying characters that exhibit either excessive uniqueness or excessive similarity in their traits is critical for maintaining overall game balance.

This paper introduces a novel clustering-based methodology to analyze character balance in Valorant. Unlike approaches focused on individual performance, our method explicitly considers team composition context. Analyzing professional match data from the Valorant Champions Tour (VCT) 2022, we use hierarchical agglomerative clustering with Jensen-Shannon Divergence (JSD) on agent co-occurrence patterns with teammates. This team-centric approach uncovers nuanced agent relationships and functional similarities missed by individual statistics or traditional role classifications, and provides a framework to track the impact of balance patches on team dynamics—crucial for iterative game design.

By focusing on professional play on a single map (Haven), we create a controlled environment to isolate the impact of team composition on agent relationships. This paper demonstrates how our clustering-based approach provides a more nuanced and insightful understanding of character balance, offering game developers a powerful tool for making informed, data-driven adjustments that promote a balanced and competitive gameplay experience. Ultimately, we aim to provide the means for fine-grained balance control by identifying the unique contributions of specific agents within the broader team context, using data collected directly from professional matches.

\section{Related Work}
Game analytics increasingly uses clustering for complex behavioral datasets to understand player behavior, optimize game design, and enhance player experience \cite{clustergame1}. Robust application requires deep understanding of both clustering algorithms and game mechanics for relevant insights \cite{clustergame1}.

Early applications of clustering in game analytics often focused on the holistic understanding of player behavior throughout entire gameplay experiences. For instance, Sifa et al. \cite{Sifa2013} employed clustering to analyze the evolving behavior of a substantial cohort of 62,000 players in Tomb Raider: Underworld. Their work demonstrated how tracking player progression and adaptation to evolving game mechanics could provide valuable feedback for game design. This approach has been extended to explore more nuanced aspects of player interaction, such as player styles and personalities. Ide and Chen \cite{Ide2023} utilized clustering to delve into player personalities and skill sets within online cooperative games like Valorant. By analyzing in-game statistics and identifying potential conflicts, their aim was to predict player compatibility and facilitate more effective team matching.

Further advancements in understanding dynamic player behavior have been achieved through the application of Bayesian clustering techniques. Drachen et al. \cite{bayesian} notably employed a semi-parametric Bayesian clustering approach, grounded in regression analysis, to analyze detailed player data. This method allowed them to identify latent player styles, focusing on in-game choices and decision-making processes rather than solely on observable actions or outcomes. A key advantage of their approach is its ability to model the fluidity of player behavior, accommodating transitions between clusters as players evolve their strategies and play styles over time.

Recent research targets computational methods for inter-character dynamics. Jaffe et al. \cite{jaffe2021} proposed a "restricted-play balance framework" using AI agents for quantitative balance evaluation via simulated gameplay. Braun et al. \cite{Braun2017} developed a clustering algorithm for Overwatch, correlating individual hero performance with win rates and segmenting players by hero-specific playstyles to find successful strategies, though focusing on individual statistics.

Our work distinguishes itself by empirically analyzing character balance from team compositions in professional play, rather than individual behavior or simulations. We analyze VCT 2022 data, using hierarchical agglomerative clustering with JSD on agent co-occurrence patterns to reveal latent functional roles beyond predefined categories. This uncovers nuanced relationships emerging from high-stakes team building and provides a quantitative framework to track how balance patches impact agent roles and team dynamics, prioritizing real-world competitive data for iterative game design.

\section{Limitations of Traditional Balance Metrics: Win Rate and Pick Rate}
While win rate and pick rate are frequently used to gauge agent strength and game balance \cite{jaffe2013understanding, jaffe2021}, they are ultimately indirect measures that often fail to fully capture the complex, skill-based dynamics of a game like Valorant. 

\begin{figure}[htbp]
\centering
\includegraphics[width=\columnwidth]{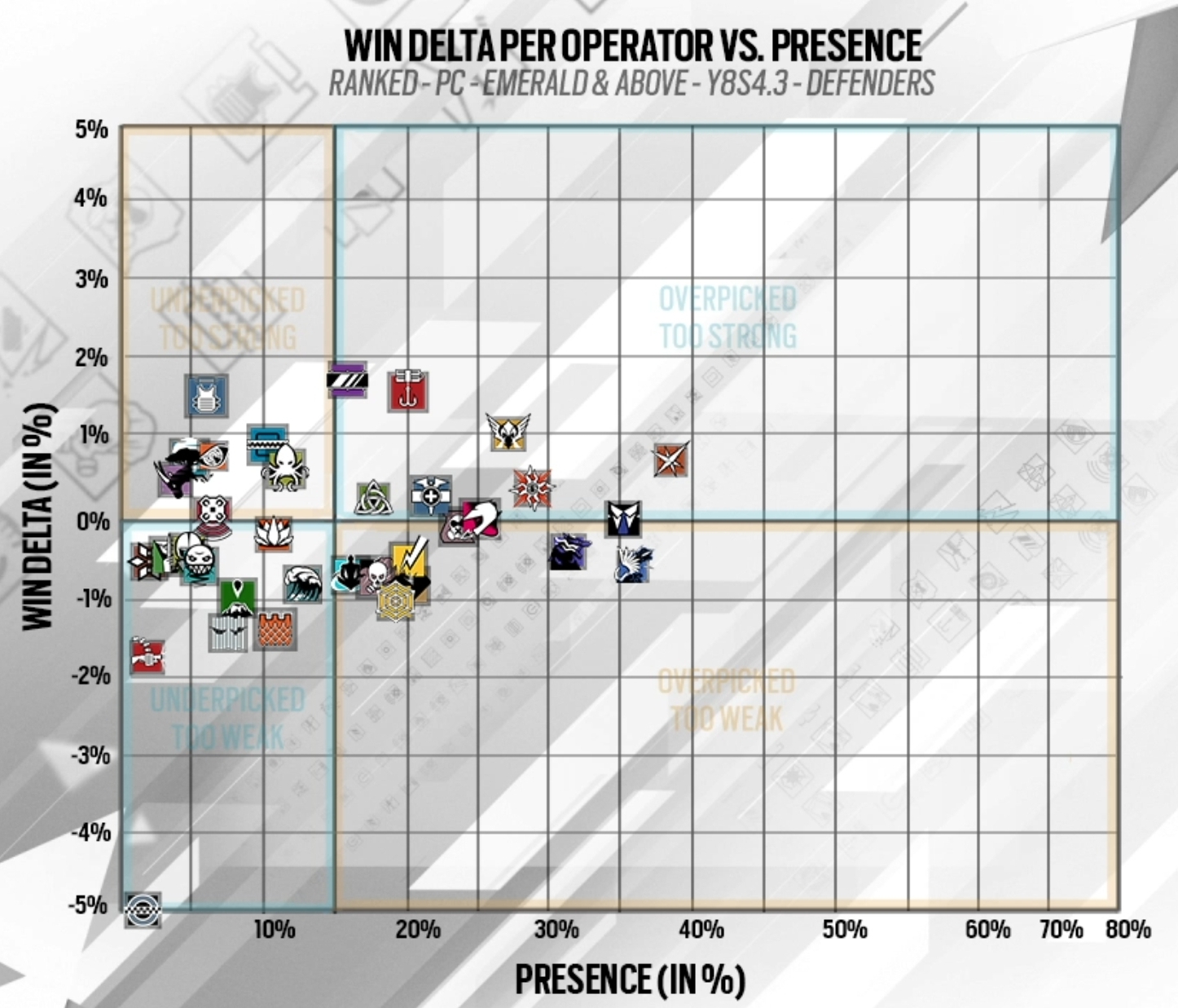}
\caption{Balancing Matrix from Ubisoft's Y9S1 Designer's Notes for Rainbow Six Siege \cite{ubisoft}. This figure plots operators based on their Win Delta (vertical axis) and Presence (horizontal axis).  The quadrants highlight operators who are overpicked/too strong, overpicked/too weak, underpicked/too strong, and underpicked/too weak.  The scatter of operators across these quadrants demonstrates the frequent lack of strong correlation between pick rate and win rate.}
\label{fig:ubisoft}  % Use a descriptive label; inside the figure environment
\end{figure}

Differing Scopes and Demonstrably Limited Correlation: Win rate measures overall success, while pick rate reflects popularity. Crucially, these two metrics are not always strongly correlated, indicating that they capture different aspects of agent performance and perceived value. As clearly demonstrated by Rainbow Six Siege's "Balancing Matrix" (Figure \ref{fig:ubisoft}) \cite{ubisoft}, their "Balancing Matrix" figures routinely show operators with high pick rates but low win deltas (and vice-versa). This lack of strong correlation is not a desirable property. It demonstrates that these metrics, while aiming for the same goal (assessing balance), are often divergent in their assessments. This divergence highlights their inadequacy as sole indicators of balance. Two benchmarks with the same goal should ideally show strong agreement; their disagreement signifies that they are missing critical information.

Oversimplification and the Need to Identify Outliers: The core weakness of win and pick rates is their inability to identify specific agents with unique, potentially balance-breaking abilities. A high win rate might indicate an overpowered agent, or it just has a high synergy with an overpowered agent. It doesn't reveal why that agent is overpowered. A low pick rate might indicate an underpowered agent, or, crucially, it might indicate an agent with a niche, highly specialized skillset that is difficult to integrate into standard team compositions. The goal of effective game balancing is often not to make all agents equally viable in all situations, but rather to ensure that no single agent or ability combination becomes overwhelmingly dominant or strategically unavoidable. Win and pick rates, being aggregate measures, mask these crucial outliers. Our methodology, in contrast, aims to expose them. By analyzing agent co-occurrence, we seek to identify agents that are infrequently chosen with a diverse range of teammates, indicating a unique, potentially disruptive role that deviates from established team composition norms. It is these outliers, not necessarily the most frequently chosen agents, that are most likely to be the source of balance problems.

\section{Data Collection and Preprocessing}

This study use Valorant as a target game, utilizes data from the Valorant Champions Tour (VCT) 2022, focusing on professional matches played on the map Haven. The dataset, sourced from the `Valorant Champion Tour 2021-2024 Data'' collection on Kaggle \cite{ref_vct_data}, was originally compiled through web scraping of vlr.gg, a prominent Valorant esports statistics website.

Our data selection is based on several justifications. First, Valorant is a highly strategic, team-based first-person shooter in which player skill sets and team compositions significantly impact game outcomes, aligning with our research objectives. VCT is Valorant's highest level of professional competition, where character selection and strategy are at their most refined. This relates to the assumption in our research methodology that professional competitions feature consistent victory goals independent of personal preferences. This consistency makes agents with similar co-occurrence patterns more aligned with the alternative hypothesis and less susceptible to noise (a more detailed description of the hypothesis is provided in Section 5.1). At the same time, limiting the time period to 2022 is an attempt to reduce the time span and lessen the impact of character skill balancing during version changes on our analysis.

Second, focusing on a single map, Haven, allows us to isolate and examine agent synergies and team compositions in a controlled environment. In Valorant, each map introduces its own strategic elements, such as specific chokepoints, sightlines, and objectives. By focusing on Haven, we minimize variability that might arise from differing map dynamics, thereby improving the precision of our analysis. The decision to work with a single map also simplifies the computational complexity, as we do not need to account for potentially diverse strategies across multiple map environments.

\subsection{Dataset Description}

The primary data source is the `teams\_picked\_agents'' file from the VCT 2022 agent data folder. This file contains detailed information about agent selections in professional Valorant matches, including variables such as Tournament, Stage, Match Type, Map, Team, Agent, Total Wins By Map, Total Loss By Map, and Total Maps Played.

\subsection{Data Preprocessing}

Key preprocessing steps included expanding Total Maps Played entries (where identical compositions were used multiple times) into individual records to maintain composition frequency integrity. We filtered for Haven matches to control for map-specific strategies. Rigorous checks ensured data integrity, including consistency of match information, five unique agents per team, and valid Agent data. A comprehensive list of unique agents was compiled for constructing characteristic vectors.

\subsection{Final Dataset}

The preprocessed dataset has 23,512 records for 24 unique agents, forming the basis for our clustering analysis of agent relationships and team strategies on Haven in VCT 2022.

\section{Method}

This study employs hierarchical agglomerative clustering combined with Jensen-Shannon Divergence (JSD) to analyze agent relationships in Valorant based on team composition tendencies. The analysis focuses on the Haven map, using data from professional matches.

\subsection{Data Representation}

In this study, we represent a given agent (e.g., agent X) by constructing its characteristic probability vector, denoted as $\mathbf{v}_X$. This vector is defined as $\mathbf{v}_X = (p_1, p_2, \dots, p_n)$, where $n$ is the total number of unique agents available in the game. Each element $p_k$ (for $k=1, \dots, n$) of the vector $\mathbf{v}_X$ corresponds to the probability that agent $k$ is selected as a teammate when agent X is picked. (Note: if agent X cannot be its own teammate, $p_X$ would be 0, or the vector could be defined over $n-1$ agents. For this study, co-occurrence implies distinct agents on a team, so the probability $p_k$ refers to agent $k$ being chosen alongside agent X). When comparing two different agents, say agent $i$ and agent $j$, their respective characteristic probability vectors will be denoted as $\mathbf{v}_i$ and $\mathbf{v}_j$. This approach is motivated by the underlying assumption that the team composition context in which an agent appears reflects its functional role in a match.

Our central hypothesis is as follows: if two different agents tend to be accompanied by largely the same set of teammates across various matches, this similarity in co-occurrence patterns can be interpreted as indicative of their substitutability. In other words, if two agents are consistently paired with nearly identical sets of teammates, they can be considered perfect substitutes in terms of their roles within the team. Consequently, the similarity between the probability vectors of two agents serves as an index of how similar (or substitutable) their roles are.

For example, if agent A’s probability vector $\mathbf{v}_A$ assigns high probabilities to the same subset of teammates as agent B’s vector $\mathbf{v}_B$, then even if A and B do not co-occur frequently, the similarity in their teammate profiles implies that they fulfill analogous roles in team compositions. Conversely, significant differences between the vectors suggest that the agents are employed in distinct tactical contexts.

By normalizing the co-occurrence counts to form these probability vectors (using the $L_1$ norm), we ensure that each vector accurately reflects the distribution of teammate selections independent of the absolute frequency of picks. This representation thus provides a robust and interpretable measure of role similarity that is used as the foundation for the subsequent distance-based clustering analysis.

\subsection{Co-occurrence Matrix Construction}

To build the co-occurrence matrix, we aggregate the occurrences of agents being selected together in teams across multiple matches. For each match, the composition of agents is noted, and we count how often each pair of agents appears in the same lineup. This co-occurrence matrix is symmetric, with each element $C_{ij}$ representing the frequency of agents $i$ and $j$ appearing together in the same team.

The co-occurrence count for each pair of agents is used to capture the relationship between agents based on their joint presence in team compositions.

\subsection{Probability Vector Normalization}

Once the co-occurrence counts are computed, we convert these counts into probability vectors for each agent. Specifically, the probability vector for each agent is obtained by normalizing the co-occurrence counts:

\[
v_{\text{norm}} = \frac{v}{\|v\|_1}
\]

where $\|v\|_1$ is the L1 norm of the vector, ensuring that each agent’s vector sums to 1. This normalization step allows for a meaningful comparison of the agents' co-occurrence patterns, accounting for varying agent pick frequencies and mitigating the impact of any agent's self-position.

\subsection{Distance Metric: Jensen-Shannon Divergence}

To measure the dissimilarity between agents, we use Jensen-Shannon Divergence (JSD) as the distance metric. JSD is a symmetric measure that quantifies the divergence between two probability distributions. Given two probability vectors $\mathbf{v}_i$ and $\mathbf{v}_j$ for agents $i$ and $j$, the JSD is computed as:

\[
d(\mathbf{v}_i, \mathbf{v}_j) = \frac{1}{2} \left( D_{\text{KL}}(\mathbf{v}_i \parallel m) + D_{\text{KL}}(\mathbf{v}_j \parallel m) \right)
\]

where $m = \frac{\mathbf{v}_i + \mathbf{v}_j}{2}$ is the mixture distribution, and $D_{\text{KL}}$ is the Kullback-Leibler divergence. JSD ensures that the distance between agents is not just based on raw co-occurrence but also on the distribution of their co-occurrences across other agents. This metric helps us capture more nuanced differences between agents' roles in team compositions.

\subsection{Clustering Algorithm}

Hierarchical agglomerative clustering (HAC) using average linkage (UPGMA), is applied to these JSD distances to reveal the hierarchical structure of agent relationships.  Average linkage is chosen for its robustness to outliers.

\subsection{Justification for Not Using Stochastic Block Model (SBM)}

We opted against using the Stochastic Block Model (SBM) because our primary goal is to understand functional agent roles and the impact of changes on those roles within team compositions, not to discover latent community structures. Our co-occurrence matrix, combined with JSD and hierarchical clustering, directly addresses this goal. This approach provides a transparent and interpretable representation of agent relationships based on observed team compositions, revealing functional similarities (as evidenced in Figure 1). Crucially, the JSD distance allows us to quantitatively track how balance patches alter agent roles and team dynamics (Section VII). This dynamic, change-focused analysis is less directly supported by SBM, which primarily focuses on static block assignments. The simplicity and directness of hierarchical clustering with JSD better serve our specific analytical needs.

\subsection{Determining the Number of Clusters} 
The selection of the number of clusters for the hierarchical agglomerative clustering was guided by statistical evaluation using the Silhouette Score. For each potential number of clusters $k$ (tested from 2 to $N-1$, where $N$ is the number of unique agents), the average Silhouette Score was calculated using the precomputed Jensen-Shannon Divergence distance matrix. Our analysis revealed that a configuration of 5 clusters yielded the maximum average Silhouette Score of 0.3246. This result, complemented by visual inspection of the dendrogram (Figure~\ref{fig2}), informed our decision to focus on 5 distinct agent clusters for analyzing team composition tendencies on the Haven map.

\subsection{Interpretation}

The resulting clusters are interpreted as groups of agents with similar roles or synergies within team compositions on the Haven map. Agents clustered together are understood to have similar roles or synergies, contributing similarly to the team's composition strategies. This approach allows us to uncover implicit agent relationships based on actual gameplay data, potentially revealing nuanced roles and synergies that may not be apparent from predefined agent categories.

By focusing on team composition tendencies, we gain deeper insights into how professional players structure their teams around different agents. This clustering methodology reveals roles that emerge dynamically from the data, capturing patterns of synergy and substitution between agents.

\section{Result}

\begin{figure}[htbp]
\centerline{\includegraphics[width=\columnwidth]{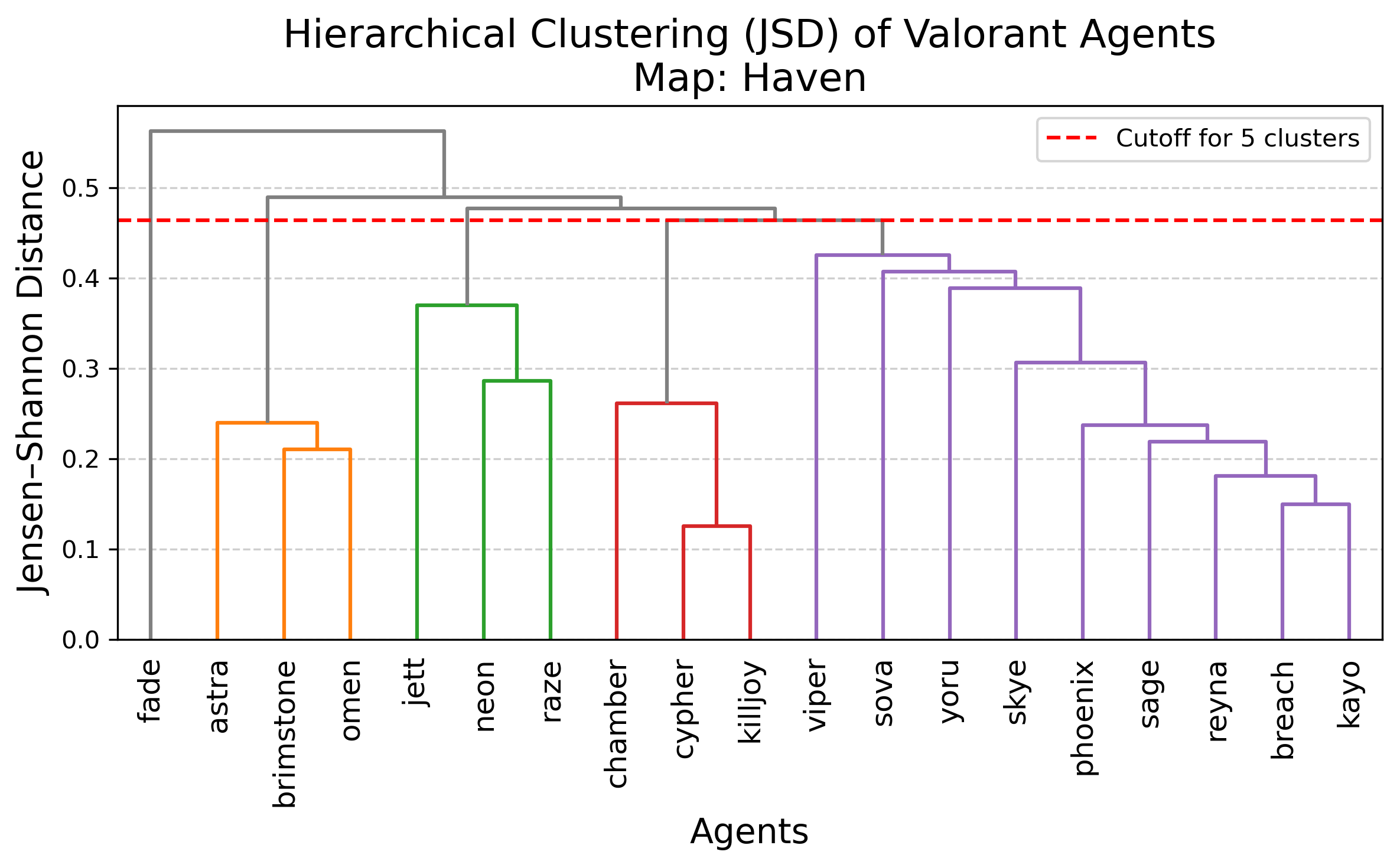}}
\caption{Hierarchical Clustering of Valorant Agents on the Haven Map. The dendrogram displays the hierarchical clustering of agents based on their team composition tendencies, measured using Jensen-Shannon Divergence. Agents with similar co-occurrence patterns (i.e., frequently paired with the same teammates) are grouped together in the same clusters. Different colors represent distinct clusters of agents with similar roles or synergies within team compositions.}
\label{fig2}
\end{figure}

Figure \ref{fig2} presents a hierarchical clustering analysis of Valorant agents on the Haven map, derived from their co-occurrence frequencies within professional team compositions. The resulting dendrogram reveals distinct clusters that largely align with, and in some cases precisely mirror, established functional roles and sub-classifications within the game. For instance, Astra, Omen, and Brimstone form a cohesive cluster, reflecting their shared designation as Controllers—agents specializing in map control and strategic utility deployment. Similarly, Chamber, Killjoy, and Cypher are grouped together, consistent with their roles as Sentinels who excel in defensive strategies, trap placement, and information gathering. The close proximity of these agents within their respective clusters in the dendrogram suggests a degree of functional interchangeability and validates these predefined role categories through empirical gameplay data.

A notable grouping comprises agents heavily reliant on flash utilities, including Phoenix, Reyna, Kayo, Skye, Yoru, and Breach. Their clustering indicates a shared functional pattern, likely stemming from their common use of flash abilities to initiate engagements and create opportunities for their team. Another highly distinct and significant cluster consists of Jett, Neon, and Raze. These agents are specifically known within the Valorant meta as 1st entry Duelists, characterized by their high mobility and aggressive kits designed for spearheading offensive plays and securing initial engagements. The clear separation of this "1st entry Duelist" group in our analysis strongly corroborates this well-understood tactical sub-classification.

The analysis also identifies Fade as a distinct outlier, whose unique skill set and impact on the Haven map position her separately from the more broadly defined role clusters. This finding aligns with our initial objective to pinpoint agents possessing unique ability profiles that deviate from standard role classifications. Such information is crucial for informing balance adjustments, allowing developers to target specific skill characteristics to maintain overall game equilibrium. The clustering methodology, therefore, not only confirms expected broad role-based groupings like Controllers and Sentinels but also successfully identifies more granular, community-recognized sub-roles such as "1st entry Duelists," and uncovers nuanced synergies. This reveals that agents' roles are not rigidly confined to predefined categories but are dynamically shaped by the interplay of their co-occurrence tendencies in competitive play.

The cluster analysis validates our approach by showing results that highly overlap with the predefined agent categorization of the Valorant development team. More importantly, it provides granular insights into agent synergies and functional similarities derived directly from professional team compositions. This data-driven perspective on agent relationships complements traditional balance metrics, offering a foundation for more nuanced understanding and context-aware balance adjustments.

\section{Measuring the Impact of Agent Adjustments on Game Balance}

Beyond clustering agents into role-based groups, the Jensen-Shannon Divergence (JSD) distances calculated between agent probability vectors offer a valuable tool for quantitatively assessing the impact of developer-initiated character adjustments on game balance. In the dynamic environment of competitive games, developers frequently release balance patches that modify agent abilities, statistics, or other attributes to refine gameplay and address balance concerns. Our methodology provides a framework to evaluate the effects of these adjustments by observing changes in agent relationships and team composition dynamics.

To measure the impact of a specific agent adjustment, this approach advocates for a comparative analysis of agent co-occurrence patterns before and after the implementation of a balance patch. Firstly, the clustering analysis, as described in Section 5, should be performed using gameplay data collected prior to the patch. This establishes a baseline understanding of agent roles and relationships within team compositions. Subsequently, after the balance patch has been deployed and sufficient gameplay data has been accumulated, the clustering analysis is repeated using the post-patch data.

The core of the impact assessment lies in comparing the JSD distances and cluster memberships of agents between the pre-patch and post-patch analyses. Several key metrics, including absolute distance changes, can be examined:

\begin{itemize}
    \item \textbf{Absolute Change in Agent Distance to Cluster Centroid ($\Delta D_{centroid}$)}: For each agent, particularly those directly affected by the balance patch, we can measure the change in its distance to the centroid of its assigned cluster. Let $C_{pre}$ be the cluster to which agent $A$ belongs in the pre-patch analysis, and let $\mu_{pre}$ be the centroid of $C_{pre}$, calculated as the average of the probability vectors of all agents in $C_{pre}$:
    \begin{equation}
        \boldsymbol{\mu}_{pre} = \frac{1}{|C_{pre}|} \sum_{\mathbf{v}_i \in C_{pre}} \mathbf{v}_i^{pre}
    \end{equation}
    where $\mathbf{v}_i^{pre}$ is the probability vector of agent $i$ pre-patch. Similarly, let $\mu_{post}$ be the centroid of the corresponding cluster $C_{post}$ in the post-patch analysis:
    \begin{equation}
        \boldsymbol{\mu}_{post} = \frac{1}{|C_{post}|} \sum_{\mathbf{v}_j \in C_{post}} \mathbf{v}_j^{post}
    \end{equation}
    Then, the absolute change in distance of agent $A$ to its cluster centroid is given by:
    \begin{equation}
        \Delta D_{centroid}(A) = |JSD(\mathbf{v}_A^{post}, \boldsymbol{\mu}_{post}) - JSD(\mathbf{v}_A^{pre}, \boldsymbol{\mu}_{pre})|
    \end{equation}
    A significant positive $\Delta D_{centroid}(A)$ might indicate a shift in agent $A$'s role away from its previous cluster archetype, potentially indicating an imbalance or role disruption. A negative or negligible change could suggest the agent's role remains consistent, or the balance adjustment has reinforced its intended role.

    \item \textbf{Absolute Change in Average Inter-Agent Distance within Clusters ($\Delta D_{inter}$)}: We can also measure how the average pairwise JSD distance between agents within a cluster changes after a balance patch. For a cluster $C$, the average inter-agent distance can be defined as:
    \begin{equation}
        \bar{D}_{inter}(C) = \frac{2}{|C|(|C|-1)} \sum_{\mathbf{v}_i \in C} \sum_{\mathbf{v}_j \in C, i<j} JSD(\mathbf{v}_i, \mathbf{v}_j)
    \end{equation}
    The absolute change in average inter-agent distance for a cluster $C$ between pre-patch and post-patch is:
    \begin{equation}
        \Delta D_{inter}(C) = |\bar{D}_{inter}(C_{post}) - \bar{D}_{inter}(C_{pre})|
    \end{equation}
    An increase in $\Delta D_{inter}(C)$ might suggest that the agents within cluster $C$ have become more differentiated in their roles or synergies post-patch, potentially indicating a destabilization within that role archetype. Conversely, a decrease could suggest increased homogeneity or clearer role definitions within the cluster.

    \item \textbf{Shift in Cluster Membership}: A more qualitative but crucial observation is whether agents change cluster membership entirely between pre-patch and post-patch analyses. This indicates a fundamental shift in an agent's role within team compositions, which can be directly observed by comparing cluster assignments in the dendrograms.

    \item \textbf{Overall Dendrogram Structure Comparison}: Visually comparing the dendrograms from pre-patch and post-patch data provides a holistic view of how balance changes affect the broader agent ecosystem. Changes in branching patterns and cluster formations can reveal systemic shifts in agent roles and team composition strategies.
\end{itemize}

\section{Conclusion}
Traditional balance metrics offer limited insight into team-based game dynamics. This paper proposed and validated a novel clustering methodology for character balance in Valorant using professional match data and team composition tendencies. Applying HAC with JSD to agent co-occurrence identified distinct agent clusters aligning with established roles and highlighting unique tactical profiles and nuanced synergies beyond predefined categories.

JSD distances also provide a quantitative framework to assess balance adjustment impacts on agent roles and team dynamics, enabling more data-driven, iterative game design. The primary contribution is a methodology offering a holistic, interpretable, context-aware perspective on character balance. By focusing on collective agent function, our approach complements existing metrics, providing developers a tool for informed adjustments promoting strategic diversity. Despite limitations, this method represents a significant step towards a more nuanced understanding of game balance.

\section{Limitation and Future Works}
Our primary assumption—that agents with similar team compositions are perfect substitutes—oversimplifies the complexities of gameplay. Agent roles are not strictly interchangeable, as players choose agents based on playstyle, synergy, and broader strategy. Two agents with similar co-occurrence data may still play distinct roles. The clustering reveals similarity in team composition, but the distance metric alone doesn't define a "balanced" vs. "imbalanced" threshold. This poses a challenge for translating the analysis into specific balance adjustments.

Future work could incorporate factors like side-specific win rates or player behavior for a more comprehensive view of balance. At the same time, developing robust metrics for determining "balance" based on distance is crucial.  While the impact of specific character skill changes on win rates varies greatly between different games and even different patches within a single game, developers can establish a relative sense of "overly large" changes. This could be achieved by normalizing the magnitude of JSD changes observed after a patch against a historical baseline of changes from previous patches.  Furthermore, correlating these quantitative JSD shifts with qualitative community feedback (e.g., forum discussions, surveys) could provide a valuable cross-validation, helping to identify changes that are perceived as disruptive by the player base, even if their direct impact on win rates is difficult to isolate. Adding subjective elements like strategy would enhance the method's utility for real-world game balancing.

\end{document}